\documentclass{article}

\usepackage{latexsym}
\usepackage{amssymb}
\usepackage{amsmath}
\usepackage{amsthm}
\usepackage{stmaryrd}
\usepackage{booktabs}
\usepackage{enumitem}
\usepackage{graphicx}
\usepackage{color}
\usepackage{tikz}
\usepackage{multirow}
\usepackage{bm}
\usepackage[utf8]{inputenc} 
\usepackage[T1]{fontenc}    
\usepackage{hyperref}       
\usepackage{url}            
\usepackage{booktabs}       
\usepackage{amsfonts}       
\usepackage{nicefrac}       
\usepackage{microtype}      
\usepackage{xcolor}         

\usepackage{algorithmic}
\usepackage{textcomp}
\usepackage{xcolor}
\usepackage[final]{neurips_2025}

\newtheorem{definition}{Definition}

\DeclareMathOperator*{\argmin}{arg\,min}

\def\BibTeX{{\rm B\kern-.05em{\sc i\kern-.025em b}\kern-.08em
    T\kern-.1667em\lower.7ex\hbox{E}\kern-.125emX}}

\def\checkmark{\tikz\fill[scale=0.23](0,.35) -- (.25,0) -- (1,.7) -- (.25,.15) -- cycle;} 

\begin{document}


\title{Natural Gradient Descent for Online Continual Learning}

\author{
  Joe Khawand \\
  Ecole Polytechnique \& Télécom Paris \\
  Paris, France \\
  \texttt{joe.khawand@polytechnique.org}
  \And
  David Colliaux \\
  Sony Computer Science Laboratories \\
  Paris, France \\
  \texttt{0000-0003-1898-4864}
}

\maketitle{}

\begin{abstract}
Online Continual Learning (OCL) for image classification represents a challenging subset of Continual Learning, focusing on classifying images from a stream without assuming data independence and identical distribution (i.i.d). The primary challenge in this context is to prevent catastrophic forgetting, where the model’s performance on previous tasks deteriorates as it learns new ones. Although various strategies have been proposed to address this issue, achieving rapid convergence remains a significant challenge in the online setting. In this work, we introduce a novel approach to training OCL models that utilizes the Natural Gradient Descent optimizer, incorporating an approximation of the Fisher Information Matrix (FIM) through Kronecker Factored Approximate Curvature (KFAC). This method demonstrates substantial improvements in performance across all OCL methods, particularly when combined with existing OCL tricks, on datasets such as Split CIFAR-100, CORE50, and Split miniImageNet.
\end{abstract}


\section{Introduction}

With the surge of embedded AI systems and the growing need for them, Continual Learning presents itself as an elegant solution to on-device learning. This paradigm refers to the ability of a model to learn by continuously assimilating knowledge over time as observed in biological entities \cite{kudithipudi_biological_2022}. Mathematically, this refers to the ability to train a model while breaking the i.i.d. assumption on the data, sending classes one after the others, for example, in the classification scenario. Naively training a model in this scenario leads to the phenomenon of \textit{catastrophic forgetting} \cite{french_catastrophic_1999,mccloskey_catastrophic_1989}, where the model forgets previously learned representations and specializes on the last seen task. 

Most of the research focuses on scenarios in which the model can iterate for multiple epochs \cite{van_de_ven_three_2022} and use batches of reasonable size. Unfortunately, doing so requires significant storage that is more often than not unavailable on devices. Online Continual Learning (OCL)  \cite{mai_online_2022,soutif--cormerais_comprehensive_2023} considers the more realistic but more difficult scenario where the model can only iterate once on the current task and has no access to the previous ones. The model is thus required to achieve satisfactory performance from a single pass over the online data stream using very small batch sizes. In this context, the model may encounter new categories (Online Class Incremental, \textbf{OCI}) or changes in the characteristics of the data such as alterations in the background, the introduction of blur or noise, changes in lighting, or the presence of obstructions (Online Domain Incremental, \textbf{ODI}) \cite{van_de_ven_three_2022}. We focus our attention on the methods that do not require access to task labels, as this does not represent a realistic OCL scenario. In this complex scenario, OCL methods struggle to achieve high accuracies as shown in the survey \cite{mai_online_2022}. There is still a big gap between OCL-trained models and Offline trained models. 

In this paper, we explore the use of 2\textsuperscript{nd} order optimization in Online Continual Learning. Motivated by the need for faster convergence, we investigate the use of the Natural Gradient Descent optimizer \cite{amari_natural_1998} with Kronecker Factored Approximate Curvature (KFAC) \cite{martens_optimizing_2020}, in the OCL scenario. We demonstrate that this method, presented as the Steepest Descent in the distribution space, boasts increased performance across all the OCL methods and tricks presented in \cite{mai_online_2022}, on datasets such as  Split CIFAR-100 \cite{krizhevsky_learning_2009}, CORE50 \cite{lomonaco_core50_2017}, and Split MiniImageNet \cite{mai_online_2022}.

\begin{figure}[t]
\begin{center}
   \includegraphics[width=\linewidth]{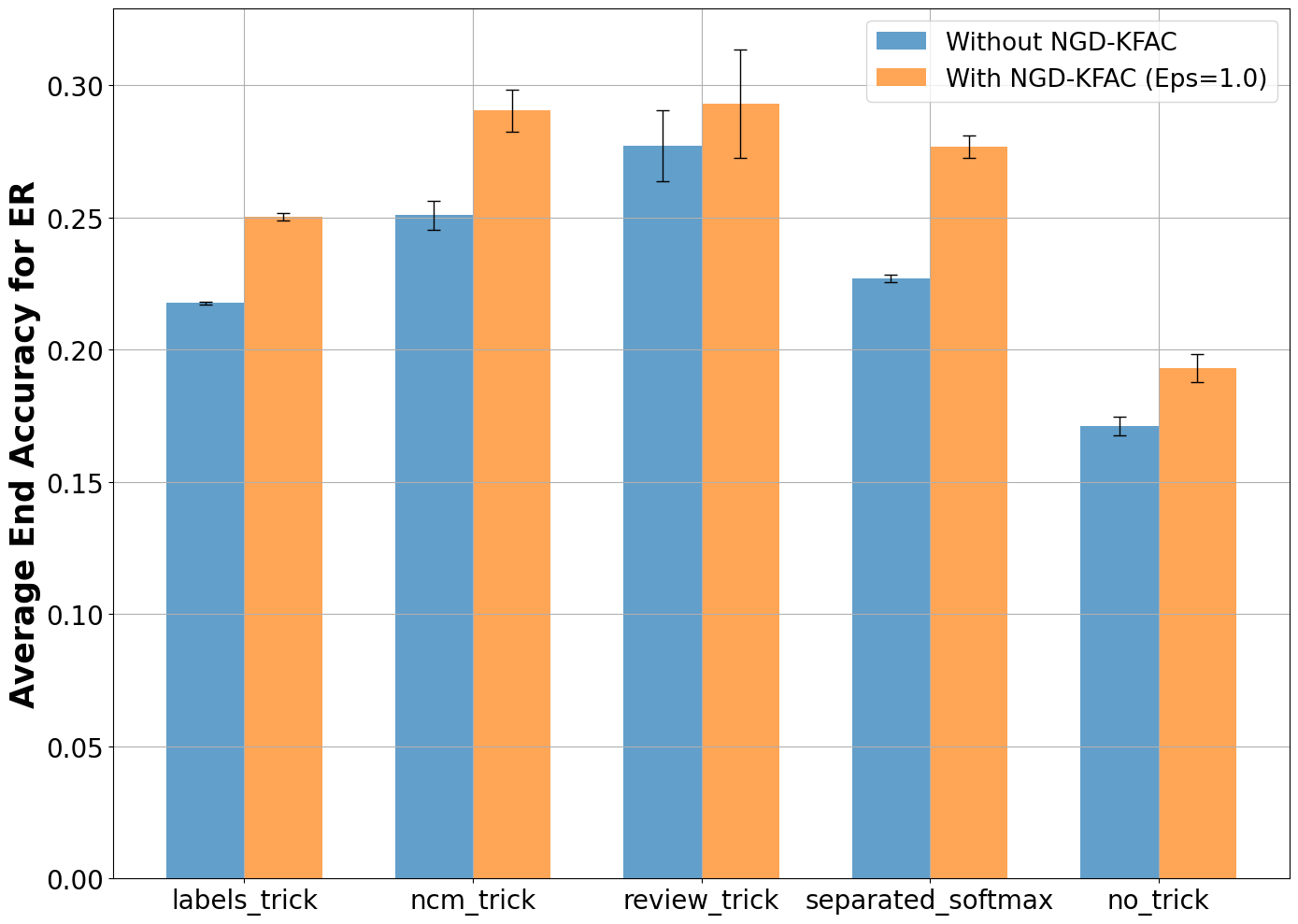}
\end{center}
   \caption{Average End Accuracy on Split-CIFAR-100 \cite{krizhevsky_learning_2009} for 3 buffer sizes (1k-5k-10k) using ER \cite{rolnick_experience_2019} in combination with OCL tricks \cite{mai_online_2022}.}
\label{fig:er_histogram_cifar}
\end{figure}

\textbf{Our paper makes the following contributions:}
\begin{enumerate}
    \item We propose the use of Natural Gradient descent with Kronecker Factored Approximate Curvature (NGD-KFAC) as an optimization method in Online Continual Learning (OCL) for image classification. This is a novel application of NGD-KFAC in the OCL domain.
    \item Our experiments show that NGD-KFAC significantly improves the performance of OCL models on various OCI and ODI benchmarks. This method enables most existing online continual learning techniques to achieve better results with only a minor change to the optimizer, highlighting its effectiveness in reducing catastrophic forgetting and enhancing model convergence.
\end{enumerate}

\section{Related work}
Continual learning has been studied to address the dynamic nature of real-world data, where the assumption of a static data distribution often does not hold \cite{van_de_ven_three_2022}. A considerable portion of this research has been dedicated to mitigating the phenomenon of catastrophic forgetting \cite{french_catastrophic_1999,mccloskey_catastrophic_1989}, a significant challenge when models are sequentially trained on multiple tasks. A comprehensive survey by Mai et al. (2022) \cite{mai_online_2022} highlights various strategies used in Online Continual Learning (OCL), showing the complexity of training models that are constrained to learn from data streams with limited passes and without access to previous tasks. This survey shows the gap between traditional offline learning and the difficult requirements of OCL scenarios.

The use of second-order optimization methods, specifically Natural Gradient Descent (NGD), has shown promise in machine learning for its ability to account for the underlying data distribution's geometry. Amari (1998) \cite{amari_natural_1998} introduced NGD, leveraging the Fisher Information Matrix (FIM) to guide optimization in a manner that respects the information geometry of parameter space, offering a more principled update path compared to first-order methods like Stochastic Gradient Descent (SGD). Despite its theoretical advantages, the practical application of NGD in deep learning has been limited by computational challenges, particularly in calculating and inverting the FIM for large models.

To overcome these challenges, Martens and Grosse (2015) \cite{martens_optimizing_2020} proposed the Kronecker-Factored Approximate Curvature (KFAC) method, an approximation of NGD that reduces the computational overhead by approximating the FIM with a Kronecker product. EKFAC, introduced by George et al. (2018) \cite{george_fast_2021}, enhances the KFAC approximation by rescaling the Kronecker factors with a diagonal matrix derived from singular value decomposition (SVD), aiming to mitigate some of these challenges. Conversely, the KBFGS \cite{goldfarb2021practical} method seeks to approximate the inverse of the Kronecker factors through low-rank Broyden-Fletcher-Goldfarb-Shanno (BFGS) updates, offering a different approach to handling the inversion problem. The TENGraD method, developed by Soori et al. \cite{soori_tengrad_2022}, is accurate, fast, and needs less memory because it inverts blocks of the Fisher Information Matrix (FIM). It does this by efficiently breaking down the FIM and reusing certain calculations. Despite the advancements offered by EKFAC, KBFGS, and TenGrad, our tests showed that KFAC was more numerically stable, especially for our specific needs. We found KFAC to be more reliable and effective in this OCL setting, which is why we preferred it.

Works \cite{zenke_continual_2017,aljundi_memory_2018,kirkpatrick_overcoming_2017,li_learning_2017} have explored the potential of second-order methods and regularization techniques to mitigate catastrophic forgetting. These methods provide a promising direction for improving model robustness and adaptability in dynamic learning environments, but they still fall short of other methods like experience replay \cite{rolnick_experience_2019}, especially in the OCL scenario\cite{mai_online_2022}. Additionally, Variational Continual Learning (VCL) \cite{nguyen2018variationalcontinuallearning, tseran2018natural} has been proposed as an improvement of those methods \cite{kirkpatrick_overcoming_2017,li_learning_2017}. However, to our knowledge, there are few direct comparisons of VCL with other methods in OCL settings, and as such, we do not include VCL in our evaluations.

This paper builds upon these works by introducing an application of NGD-KFAC, within the OCL framework. Our approach leverages the strengths of second-order optimization to address the challenges of OCL, such as rapid convergence and efficient learning with data non-stationarity and task shifts. By integrating NGD with KFAC, our aim is to bridge the gap identified by Mai et al. (2022) \cite{mai_online_2022}, enhancing the performance of OCL models on benchmarks such as Split CIFAR-100 \cite{krizhevsky_learning_2009}, CORE50 \cite{lomonaco_core50_2017}, and Split MiniImageNet \cite{mai_online_2022}.

\section{Continual Learning}

Continual Learning (CL) addresses the challenge of training models on non-stationary data distributions, where the data \( \mathcal{D} = \bigcup_{t=1}^T \mathcal{D}_t \) is presented as sequential tasks \( \mathcal{D}_t \), each drawn from a distinct distribution \( p_t(X, y) \). In this setting, the i.i.d. assumption is violated, and training naively on \( \mathcal{D}_t \) results in catastrophic forgetting \cite{french_catastrophic_1999}, where model parameters \( \theta \) are optimized for \( \mathcal{D}_t \) at the cost of degrading performance on \( \mathcal{D}_1, \dots, \mathcal{D}_{t-1} \).

Mitigating forgetting has been approached using regularization methods that constrain updates to \( \theta \) based on parameter importance \cite{kirkpatrick_overcoming_2017}, experience replay, which stores subsets of previous data to approximate \( p(X, y) \) \cite{rolnick_experience_2019}, or architectural growth to isolate task-specific parameters \cite{rusu2022progressiveneuralnetworks}. However, these methods often assume multiple passes over data or large storage, making them unsuitable for online learning scenarios.

\section{Online Continual Learning}

Online Continual Learning (OCL) extends CL by imposing stricter constraints: each data point is processed only once (\( T=1 \)) and prior tasks \( \mathcal{D}_{1:t-1} \) are inaccessible. Formally, at each step \( t \), the model minimizes a streaming loss \( \mathcal{L}_t(\theta) = \mathbb{E}_{(X, y) \sim p_t}[\ell(f_\theta(X), y)] \) using small batch updates while adapting to the evolving \( p_t \) \cite{mai_online_2022}. 

OCL can be divided into Online Class Incremental (OCI), where new classes are introduced over time, and Online Domain Incremental (ODI), where \( p_t(X, y) \) shifts without introducing new classes \cite{van_de_ven_three_2022}. Without task labels or memory buffers, the challenge is to update \( \theta \) efficiently while minimizing catastrophic forgetting and adapting to non-stationary data streams. This paper explores using Natural Gradient Descent (NGD) with Kronecker-Factored Approximate Curvature (KFAC) to improve convergence and robustness in these constrained settings.

\section{Natural Gradient Descent}

Traditional Stochastic Gradient Descent (SGD) relies solely on the gradient of the loss function \( \mathcal{L} \), as defined by the update rule:

\begin{equation}
\theta_{t+1} = \theta_{t} - \alpha \nabla_{\theta_{t}} \mathcal{L}(\theta_{t} | X)
\end{equation}

where \( \alpha \) is the learning rate. This method chooses the direction in the \( \theta \) space so that the loss \( \mathcal{L} \) decreases the most i.e. highest reduction in loss with a unit change in the parameter \( \theta \).


However, the Euclidean metric underlying this method may not always provide the most efficient path for updating parameters. Indeed, a slight adjustment in the parameter space can lead to a disproportionate change in the distribution. The conventional gradient descent method fails to account for this by not considering the curvature of the distribution space.

To address those shortcomings, NGD proposes stepping in the direction that better respects the geometry of the distribution space \cite{amari_natural_1998, shrestha_natural_2023}. It does so by employing the Fisher Information Matrix (FIM) to adjust the gradient, which accounts for the curvature of the parameter space as informed by the data distribution. Our intuition, for using this optimizer in \textbf{Online Continual Learning} stems from the need to achieve faster convergence, and avoid huge perturbation due to the non i.i.d constraint set on the dataset.

The update rule for Natural Gradient Descent can be expressed as:

\begin{equation}
    \theta_{t+1} = \theta_{t} - \alpha \cdot F^{-1}(\theta_{t}) \cdot \nabla \mathcal{L}(\theta_{t})
\end{equation}

where \(F^{-1}(\theta_{t})\) is the inverse of the Fisher Information Matrix (FIM) and \(\nabla \mathcal{L}(\theta_{t})\) is the gradient of the loss function, where the FIM is defined as follows.

\begin{definition}
The Fisher Information Matrix is defined as:
\begin{equation}
        F(\theta_{t})=\mathbb{E}_{(x,y)\in\mathcal{D}_{\text{train}}}[\frac{\partial \ell (f_{\theta}(x),y)}{\partial \theta}\frac{\partial \ell (f_{\theta}(x),y)}{\partial \theta}^{T}]
\end{equation}
where the expectation is taken over targets sampled from the model $p_{\theta} = f_{\theta}$
\end{definition}

But as it is formulated, the matrix $F$ has a gigantic size $n_{\theta} \times n_{\theta}$ (with $n_{\theta}$ the number of parameters) which makes it too large to compute and invert in the context of modern deep neural networks with millions of parameters.

That is why research in this domain gravitated towards approximation methods \cite{heskes_natural_2000,desjardins_natural_2015,martens_optimizing_2020,fujimoto_neural_2017,ba_distributed_2016} for the Fisher information matrix.

\subsection{Kronecker Factored Curvature Approximation}
Using the exact FIM, the storage required is proportional to the square of the number of parameters and the computation of the inverse of FIM is proportional to its cube. To mitigate this problem, most of the natural gradient methods approximate FIM using a block-diagonal matrix such that the elements corresponding to cross layers are 0 \cite{shrestha_natural_2023}. 

But even with this approximation storage and computation requirements are still high. Martens and Grosse \cite{martens_optimizing_2020} mitigate this by approximating the FIM by the Kronecker product of two small matrices.

Considering the parameters $\theta_{l}$ of a layer $l$ of size $(n,m)$, $h$ the input of that layer and $g=\frac{\partial \mathcal{L}}{\partial \theta_{l}}$, the KFAC approximation of the FIM part associated to layer $l$, $F^{(l)}$, is approximated as follows  \cite{martens_optimizing_2020}:

\begin{equation}
    F^{(l)} \approx A \otimes B
\end{equation}
where :
\begin{align}
    A_{i,i'}&= \mathbb{E}[h_{i}h_{i'}] , \quad \forall (i,i') \in \llbracket 1,n\rrbracket ^{2}  \\
    B_{j,j'}&= \mathbb{E}[g_{j}g_{j'}] , \quad \forall (j,j') \in \llbracket 1,m\rrbracket ^{2}
\end{align}

\subsection{Regularization and numerical stability}
2\textsuperscript{nd} order optimization problems can be understood using a more traditional optimization perspective as presented by Martens \cite{martens_new_2014}. The idea is to compute the update $\delta$ to $\theta \in \mathbb{R}^{n}$ by minimizing a local quadratic approximation (or "model" $M$) of the loss function centered around the current parameters at this iteration. We thus consider the model $M$:

\begin{definition}
    \begin{equation}
        M_k(\delta) = \frac{1}{2} \delta^T F_k \delta + \nabla l(\theta_k)^T \delta + \ell(\theta_k),
    \end{equation}
    where:
    \begin{itemize}
        \item $M_k$ is the model at iteration k.
        \item $F_k \in \mathbb{R}^{n_{\theta} \times n_{\theta}}$ is the curvature matrix which corresponds to the FIM in our case.
        \item $\ell$ is our objective function, which consists of minimizing the average loss. This is defined by: \begin{equation}
            \ell(\theta) = \frac{1}{|\mathcal{X}|} \sum_{(x,y) \in (\mathcal{X},\mathcal{Y})} \mathcal{L}(y, f(x, \theta))
        \end{equation}
        \item $\delta$ is the update to $\theta$
    \end{itemize}
\end{definition}

With this definition, the natural gradient descent becomes the minimizer of $M$ which is a convex approximation of the 2nd-order Taylor series of expansion of $l(\delta + \theta)$. 

We can thus notice that this optimization method succeeds in generating a good local update as long as $M(\delta)$ is a good local approximation of $l(\delta+\theta)$. This is why \cite{martens_new_2014,martens_optimizing_2020} argue that it is necessary to use damping techniques such as the popular Tikhonov regularization \cite{More1977TheLA} to prevent breakdowns in local quadratic approximations. These breakdowns usually tend to occur at the beginning of training since at first, the model tends to experience an initial "exploration phase" \cite{Darken1990NoteOL}. \textbf{This is especially relevant in our case since in OCL, the model is always stuck in this initial exploration phase thus requiring heavy damping.} We showcase in section \ref{sec:res} that, in our experiments, using a high damping of $1.0$ produces the best results across all datasets and techniques.

For our experiments, we use the adapted Tikhonov regularization presented by \cite{martens_optimizing_2020}. For a layer $l \in \mathbb{N}$ with $d_{l} \in \mathbb{N}$ units we have:

\begin{equation}
    F^{(l)}=\left( A_{l-1,l-1} + \pi_l \left( \sqrt{\lambda} I \right) \right) \otimes \left( B_{l,l} + \frac{1}{\pi_l} \left( \sqrt{\lambda} I \right) \right)
\end{equation}

Where $\lambda \in \mathbb{R}^+$ is our damping parameter and $\pi$ is defined by:

\begin{equation}
    \pi_l = \sqrt{ \frac{\text{tr}(A_{l-1,l-1})/(d_{l-1} + 1)}{\text{tr}(B_{l,l})/d_l} }
\end{equation}

\section{Experimental Setting}

We consider sets $\mathcal{X}$ and $\mathcal{Y}$, representing our datapoints and labels, respectively. We aim to train a model $M$ with parameters $\theta$ to accurately classify classes in  $\llbracket1, C\rrbracket$, where $C$ is a positive integer. This is done in an Online fashion where the model can only iterate for one epoch on the data.

\subsection{Datasets}
We use the same datasets as the OCL survey \cite{mai_online_2022}:
\begin{enumerate}
    \item \textbf{Split CIFAR-100:} CIFAR100 \cite{krizhevsky_learning_2009} divided into 20 tasks with disjoint classes, each containing 5 classes.
    \item \textbf{Split MiniImageNet:} Splits the MiniImageNet \cite{vinyals2017matching} dataset with 100 classes, into 20 disjoint tasks as in \cite{chaudhry2019tiny} containing each 5 classes.
    \item \textbf{NonStationary-MiniImageNet:} Proposed by \cite{mai_online_2022} this dataset uses 3 nonstationary domain incremental scenarios: noise, blur, and occlusion. In our experiment, this is divided into 10 tasks for each type.
    \item \textbf{CORe50-NC:} \cite{lomonaco_core50_2017} is a benchmark designed for class incremental learning with 9 tasks and 50 classes: 10 classes in the first task and 5 classes in the subsequent 8 tasks.
    \item \textbf{CORe50-NI:} \cite{lomonaco_core50_2017} is a benchmark designed for assessing the domain incremental learning with 8 tasks, where each task has 50 classes with different types of nonstationarity including illumination, background, occlusion, pose and scale.
\end{enumerate}

\begin{table}[h]
\centering
\caption{Summary of the datasets used.}
\begin{tabular}{l|cccccc}
\toprule
Dataset & \#Task & \#Train/task & \#Class & Setting \\
\midrule
Split CIFAR-100 \cite{krizhevsky_learning_2009} & 20 & 2500  & 100  & OCI \\
Split MiniImageNet \cite{chaudhry2019tiny} & 20 & 2500  & 100 & OCI \\
NS-MiniImageNet \cite{mai_online_2022} & 10 & 5000  & 100 & ODI \\
CORE50-NC \cite{lomonaco_core50_2017} & 9 & 12000$\sim$24000 & 50 & OCI \\
CORE50-NI \cite{lomonaco_core50_2017} & 8 & 15000 & 50  & ODI \\
\bottomrule
\end{tabular}
\end{table}

For the NonStationary-MiniImageNet dataset \cite{mai_online_2022}, we use the same strengths of nonstationarity as in \cite{mai_online_2022} to be able to better compare the results. The nonstationarity strength increases as the experiments goes, testing the model's capability to adapt to noisier domains from the previous learning experience. Here are the values we used:
\begin{itemize}
    \item Noise: $[0.0, 0.4, 0.8, 1.2, 1.6, 2.0, 2.4, 2.8, 3.2, 3.6]$
    \item Occlusion: $[0.0, 0.07, 0.13, 0.2, 0.27, 0.33, 0.4, 0.47, 0.53, 0.6]$
    \item Blur: $[0.0, 0.28, 0.56, 0.83, 1.11, 1.39, 1.67, 1.94, 2.22, 2.5]$
\end{itemize}

\subsection{Models}
Using a Resnet18 \cite{he2015deep} we explore the use of different CL techniques in combination with the NGD-KFAC optimizer. We mainly focus on techniques using experience replay \cite{rolnick_experience_2019} as it has been shown \cite{mai_online_2022,soutif--cormerais_comprehensive_2023} that they offer the best performance in this OCL scenario:
\begin{itemize}
    \item \textbf{ER \cite{rolnick_experience_2019}:} Experience replay, one of the most popular CL techniques, proposes the use of a buffer to store previously seen experiences and reintegrate them into future training tasks.
    \item \textbf{A-GEM \cite{hu2020gradient,lopez-paz_gradient_2022}:} Similar to ER, A-GEM  uses a buffer to store previously encountered datapoints but adds an optimization step ensuring that the average loss for all past tasks does not increase.
    \item \textbf{MIR \cite{aljundi_online_2019}:} MIR is another replay technique that focuses on improving the buffer retrieval strategy. Where this is done randomly in ER, MIR chooses replay samples that will be the most affected by the upcoming parameter update, i.e. the samples for which the loss will increase the most after the update.
    \item \textbf{GSS \cite{aljundi2019gradient}:} Similar to MIR, GSS tries to improve the buffer retrieval strategy by diversifying the gradient directions of the samples stored in the buffer. It does so by calculating a score for each sample in the buffer given by the maximal cosine similarity between the gradient of the sample and the gradients of a random subset of the buffer. We will use this technique only for the domain incremental or new instances settings, as it has been shown in this survey \cite{mai_online_2022} that this method falls short of the others shown here.
    \item \textbf{Finetune:} Naively trains a model with no CL technique. This serves as a lower bound for our experiments.
    \item \textbf{Offline:} We train the model offline for 70 epochs with a batch size of 128.
\end{itemize}

We run our tests using a Resnet18 \cite{he2015deep} with a learning rate of $0.1$, and a fixed batch size of $10$ to mimic a realistic Online Learning Scenario. Although it has been shown that it offers better performance \cite{lomonaco2020cvpr}, we chose not to use pre-trained weights in our experiments to maintain consistency with the methodologies of the referenced papers.

We use SGD without any parameters and add NGD-KFAC with a running average of 0.9 and a damping of 1.0. We do not use ADAM \cite{kingma2017adam} for comparison because it yields worse results than SGD on this specific case of OCL with or without NGD-KFAC.

\subsubsection{OCL tricks}

Recent works \cite{masana_class-incremental_2022,hou_learning_2019,wu_large_2019,ahn_ss-il_2022,mai_online_2022} have shown that the Softmax layer and its associated Fully-Connected layer suffer from \textit{task-recency bias}, where those layers tend to be biased to the last encountered classes. This has prompted the creation of multiple tricks to alleviate this problem:

\begin{itemize}
    \item \textbf{Labels Trick} \cite{zeno_task_2019}: Cross-entropy loss calculation considers only the classes present in the mini-batch, preventing excessive penalization of logits for classes absent from the mini-batch.
    The loss function is given as:
    \begin{equation}
        \mathcal{L}_{\text{CE}}(x_i, y_i) = - \log \frac{e^{s_{y_i}}}{\sum_{j \in C_{\text{cur}}} e^{s_j}}
    \end{equation}
    where \( C_{\text{cur}} \) denotes the classes in the current mini-batch. 

    
    \item \textbf{Nearest Class Mean Classifier}: Replaces the last biased fully connected classification layer by a nearest mean classifier such as in iCarl \cite{rebuffi_icarl_2017}.
    The prototype vector for each class is given by:
    \begin{equation}
        \mu_y = \frac{1}{|M_y|} \sum_{x_m \in M_y} \phi(x_m)
    \end{equation}
    and the class label is assigned by:
    \begin{equation}
    y^* = \argmin_{y=1,...,l} \Vert \phi(x) - \mu_y \Vert
    \end{equation}
    \item \textbf{Separated Softmax} \cite{ahn_ss-il_2022}: Since one softmax layer results in a bias explained in \cite{masana_class-incremental_2022,hou_learning_2019,wu_large_2019,ahn_ss-il_2022,mai_online_2022}, this technique employs two Softmax layers one for old classes and one for new classes. Thus training new classes will not overly penalize the old logits.
    The loss function can be calculated as below:
    \begin{equation}
    \begin{split}
        \mathcal{L}(x_i, y_i) = - \log \frac{e^{s_{y_i}}}{\sum_{j \in C_{\text{old}}} e^{s_j}} \cdot 1\{y_i \in C_{\text{old}}\} -\\ \log \frac{e^{s_{y_i}}}{\sum_{j   \in C_{\text{new}}} e^{s_j}} \cdot 1\{y_i \in C_{\text{new}}\}
    \end{split}
    \end{equation}
    \item \textbf{Review trick} \cite{ferrari_end--end_2018}: Adds an additional fine-tuning step using a balanced subset of the memory buffer.
\end{itemize}
\subsection{Metrics}
We evaluate our models using two different popular metrics: 
\begin{itemize}
    \item \textbf{Average accuracy:} Average Accuracy is defined as 
        \begin{equation}
    \label{avgacc}
        A_i = \frac{1}{i} \sum_{j=1}^{i} a_{i,j}
    \end{equation}
   where \( i = T \), \( A_T \) represents the average accuracy by the end of training with the whole data sequence.

    \item \textbf{Average forgetting:} Average Forgetting at task \( i \) is defined as 
    \begin{equation}
    \label{avgfgt}
        F_i = \frac{1}{i - 1} \sum_{j=1}^{i-1} f_{i,j}
    \end{equation}
    where \( f_{k,j} = \max_{\ell \in \{1, \ldots, k-1\}} (a_{\ell,j}) - a_{k,j}, \ \forall j < k \). \( f_{i,j} \) represents how much the model has forgot about task \( j \) after being trained on task \( i \). Specifically, \( \max_{\ell \in \{1, \ldots, k-1\}} (a_{\ell,j}) \) denotes the best test accuracy the model has ever achieved on task \( j \) before learning task \( k \), and \( a_{k,j} \) is the test accuracy on task \( j \) after learning task \( k \).
\end{itemize}

We run every experiment 10 times and provide the mean and standard deviation for each experiment. For that, we use the same sequence of seeds for every model. In our case, the seed not only influences the initialisation parameters and the overall training of our model but also the order and organization of the datasets especially in the class incremental scenario,

\section{Results \& Discussion}\label{sec:res}
We notice an improvement in accuracy on most methods when we use NGD-KFAC. The results are provided in tables 2 to 5 and further discussed in this section for Online Class Incremental (OCI) learning and Online Domain Incremental (ODI) learning.

\begin{table*}[t]
\centering
\caption{End Average Accuracy of methods and tricks with or without NGD-KFAC for the OCI setting on Split CIFAR-100.}
\label{table:cifar100}
\resizebox{\textwidth}{!}{%
\begin{tabular}{@{}l|c|ccc|ccc|ccc@{}}
\toprule
&Finetune& & \multicolumn{7}{c}{$3.7 \pm 0.3$} &\\
&Offline&  & \multicolumn{7}{c}{$49.7 \pm 2.6$} & \\
\midrule
\multirow{2.5}{*}{Trick} & \multirow{2.5}{*}{NGD-KFAC} & \multicolumn{3}{c}{A-GEM} & \multicolumn{3}{c}{ER} & \multicolumn{3}{c}{MIR} \\ 
\cmidrule(lr){3-11}
 & & M=1k & M=5k  & M=10k & M=1k & M=5k & M=10k & M=1k & M=5k & M=10k \\ 
\midrule
\multirow{2}{*}{N/A} & \texttimes & $5.6 \pm 0.4$ & $5.8 \pm 0.4$ & $5.6 \pm 0.4$ & $11.1 \pm 0.7$ & $21.4 \pm 1.2$ & $21.8 \pm 1.0$ & $\bm{11.5 \pm 0.6}$ & $21.8 \pm 0.8$ & $24.7 \pm 1.3$ \\
  & \checkmark & $\bm{6.7 \pm 0.4}$ & $\bm{6.9 \pm 0.2}$ & $\bm{6.9 \pm 0.3}$ & $\bm{11.4 \pm 0.4}$ & $\bm{22.9 \pm 1.1}$ & $\bm{23.9 \pm 0.8}$ & $11.2 \pm 0.7$ & $\bm{23.1 \pm 0.7}$ & $\bm{25.9 \pm 0.8}$ \\
\midrule
\multirow{2}{*}{LB \cite{zeno_task_2019}} & \texttimes & $8.5 \pm 0.5$ & $8.5 \pm 0.6$ & $8.1 \pm 0.6$ & $17.0 \pm 0.9$ & $21.2 \pm 1.0$ & $22.5 \pm 0.7$ & $17.6 \pm 0.6$ & $22.4 \pm 0.9$ & $22.9 \pm 0.9$ \\
  & \checkmark & $\bm{10.7 \pm 0.6}$ & $\bm{11.3 \pm 0.7}$ & $\bm{10.4 \pm 0.8}$ & $\bm{19.4 \pm 1.1}$ & $\bm{25.6 \pm 0.6}$ & $\bm{27.3 \pm 0.5}$ & $\bm{19.2 \pm 0.7}$ & $\bm{25.2 \pm 0.7}$ & $\bm{27.2 \pm 0.7}$ \\
\midrule
\multirow{2}{*}{SS \cite{ahn_ss-il_2022}} & \texttimes & $8.3 \pm 0.5$ & $8.7 \pm 0.5$ & $8.6 \pm 0.4$ & $16.9 \pm 0.8$ & $23.8 \pm 1.2$ & $25.5 \pm 1.1$ & $17.4 \pm 0.8$ & $25.0 \pm 0.9$ & $26.8 \pm 1.1$ \\
  & \checkmark & $\bm{11.9 \pm 0.9}$ & $\bm{12.4 \pm 0.7}$ & $\bm{11.9 \pm 0.7}$ & $\bm{20.3 \pm 0.5}$ & $\bm{28.8 \pm 0.7}$ & $\bm{33.0 \pm 0.6}$ & $\bm{20.3 \pm 0.6}$ & $\bm{28.4 \pm 0.8}$ & $\bm{32.6 \pm 0.8}$ \\
\midrule
\multirow{2}{*}{RV \cite{ferrari_end--end_2018}} & \texttimes & $6.6 \pm 0.4$ & $\bm{15.7 \pm 1.5}$ & $\bm{24.2 \pm 1.3}$ & $\bm{14.6 \pm 0.4}$ & $31.2 \pm 1.0$ & $36.8 \pm 1.1$ & $\bm{13.2 \pm 0.6}$ & $30.8 \pm 0.8$ & $38.1 \pm 1.2$ \\
  & \checkmark & $\bm{6.9 \pm 0.3}$ & $10.7 \pm 0.7$ & $22.4 \pm 1.0$ & $13.7 \pm 0.5$ & $\bm{34.2 \pm 0.3}$ & $\bm{39.9 \pm 0.5}$ & $12.3 \pm 0.5$ & $\bm{31.7 \pm 0.7}$ & $\bm{39.6 \pm 0.5}$ \\
\midrule
\multirow{2}{*}{NCM \cite{rebuffi_icarl_2017}} & \texttimes & $11.6 \pm 0.6$ & $13.8 \pm 0.7$ & $14.6 \pm 0.6$ & $16.9 \pm 0.6$ & $28.1 \pm 1.0$ & $30.1 \pm 1.0$ & $16.6 \pm 0.5$ & $27.6 \pm 0.7$ & $31.1 \pm 1.0$ \\
  & \checkmark & $\bm{15.3 \pm 0.7}$ & $\bm{18.3 \pm 0.6}$ & $\bm{18.7 \pm 0.7}$ & $\bm{19.3 \pm 0.8}$ & $\bm{31.9 \pm 0.9}$ & $\bm{34.3 \pm 0.7}$ & $\bm{17.7 \pm 0.7}$ & $\bm{30.6 \pm 0.6}$ & $\bm{35.0 \pm 0.6}$ \\

\bottomrule
\end{tabular}%
}
\end{table*}

\begin{table*}[h]
\centering
\caption{ Average End Accuracy for NonStationary-MiniImageNet (Noise, Occlusion, Blur) and CORe50-NI in the ODI setting.}
\label{tab:average_accuracy}
\resizebox{\textwidth}{!}{%
\begin{tabular}{@{}l|c|ccc|ccc|ccc|ccc@{}}
\toprule
Method &NGD-KFAC& \multicolumn{3}{c|}{Mini-ImageNet-Noise} & \multicolumn{3}{c|}{Mini-ImageNet-Occlusion} & \multicolumn{3}{c|}{Mini-ImageNet-Blur} & \multicolumn{3}{c}{CORe50-NI} \\ \midrule
Finetune &\texttimes& \multicolumn{3}{c|}{$11.1 \pm 1.0$ } & \multicolumn{3}{c|}{$13.8 \pm 1.6$}& \multicolumn{3}{c|}{$2.4\pm 0.2 $} & \multicolumn{3}{c}{$14.0 \pm 2.8 $}\\
Offline  &\texttimes& \multicolumn{3}{c|}{$37.3 \pm  0.8$ }& \multicolumn{3}{c|}{$38.6 \pm 4.7$}& \multicolumn{3}{c|}{$11.9\pm 1.0 $}& \multicolumn{3}{c}{$51.7 \pm 1.8 $} \\
\midrule
 &Buffer Size & M=1k & M=5k & M=10k & M=1k & M=5k & M=10k & M=1k & M=5k & M=10k & M=1k & M=5k & M=10k \\
 \midrule
\multirow{2}{*}{ER} &\texttimes& 19.4 ± 1.3 & 21.6 ± 1.1 & 24.3 ± 1.2 & 19.2 ± 1.5 & 23.4 ± 1.4 & 23.7 ± 1.1 & 5.3 ± 0.6 & \textbf{8.6 ± 0.8} & 9.4 ± 0.7 & 24.1 ± 4.2 & 28.3 ± 3.5 & 30.0 ± 2.8 \\
           &\checkmark& \textbf{21.0 ± 0.9} & \textbf{25.1 ± 0.9} & \textbf{26.7 ± 0.6}  &  \textbf{21.0 ± 0.9} & \textbf{25.1 ± 0.9} & \textbf{26.7 ± 0.6}  & 5.3 ± 0.5 & 7.9 ± 0.7 & \textbf{10.0 ± 0.7}  &   \textbf{30.5 ± 1.0} & \textbf{32.5 ± 2.2} & \textbf{33.8 ± 1.6} \\ 
 \midrule
\multirow{2}{*}{MIR} &\texttimes& 18.1 ± 1.1 & \textbf{22.5 ± 1.4} & 24.4 ± 0.9 & 17.6 ± 0.7 & 22.0 ± 1.1 & 23.8 ± 1.2 & 5.5 ± 0.5 & \textbf{8.1 ± 0.6} & \textbf{9.6 ± 1.0} & 26.5 ± 1.0 & 34.0 ± 1.0 & 33.3 ± 1.7 \\
            &\checkmark& \textbf{19.4 ± 1.0} & 22.4 ± 1.1 & \textbf{25.6 ± 1.1} & \textbf{21.5 ± 1.0} & \textbf{23.7 ± 1.1} & \textbf{26.2 ± 1.0} & \textbf{5.6 ± 0.6} & 7.6 ± 0.5 & 8.6 ± 0.5 & \textbf{30.3 ± 1.9} & \textbf{35.2 ± 1.4} & \textbf{36.8 ± 1.3}\\ \midrule
\multirow{2}{*}{GSS} &\texttimes& 18.9 ± 0.8 & 21.4 ± 0.9 & 23.2 ± 1.1 & 17.7 ± 0.8 & 21.0 ± 2.2 & 23.2 ± 1.4 & 5.2 ± 0.5 & \textbf{7.6 ± 0.6} & \textbf{8.0 ± 0.6} & 25.5 ± 2.1 & 27.2 ± 2.0 & 25.3 ± 2.1 \\
            &\checkmark& \textbf{20.9 ± 0.7} & \textbf{24.4 ± 0.9} & \textbf{24.9 ± 1.0} & \textbf{21.2 ± 0.9} & \textbf{24.4 ± 0.8} & \textbf{26.8 ± 0.9} & \textbf{5.6 ± 0.7} & 7.2 ± 0.7 & 7.8 ± 0.7 & \textbf{29.0 ± 1.6} & \textbf{29.5 ± 2.4} & \textbf{29.3 ± 1.8} \\ \midrule
\multirow{2}{*}{A-GEM} &\texttimes& 14.0 ± 1.3 & 14.6 ± 0.7 & 14.2 ± 1.4 & 16.4 ± 0.7 & 13.9 ± 2.6 & 14.4 ± 2.0 & 4.4 ± 0.4 & 4.4 ± 0.4 & 4.3 ± 0.5 & 12.4 ± 1.1 & 13.8 ± 1.2 & 15.0 ± 2.2 \\
              &\checkmark& \textbf{17.8 ± 0.6} & \textbf{18.5 ± 0.7} & \textbf{18.3 ± 0.6}  &  \textbf{19.2 ± 1.3} & \textbf{19.2 ± 0.9} & \textbf{19.6 ± 1.4} & \textbf{4.7 ± 0.4} & \textbf{4.8 ± 0.3} & \textbf{4.4 ± 0.5}  & \textbf{19.4 ± 2.8} & \textbf{19.4 ± 2.2} & \textbf{19.3 ± 2.6} \\
\bottomrule
\end{tabular}%
}
\end{table*}

\begin{table*}[t]
\centering
\caption{End Average Accuracy of methods and tricks with or without NGD-KFAC for the OCI setting on Split MiniImageNet.}
\label{table:mini_imagenet}
\resizebox{\textwidth}{!}{%
\begin{tabular}{@{}l|c|ccc|ccc|ccc@{}}
\toprule
&Finetune& & \multicolumn{7}{c}{$3.4 \pm 0.2$} &\\
&Offline&  & \multicolumn{7}{c}{$51.9 \pm 0.5$} & \\
\midrule
\multirow{2.5}{*}{Trick} & \multirow{2.5}{*}{NGD-KFAC} & \multicolumn{3}{c}{A-GEM} & \multicolumn{3}{c}{ER} & \multicolumn{3}{c}{MIR} \\ 
\cmidrule(lr){3-11}
 & & M=1k & M=5k  & M=10k & M=1k & M=5k & M=10k & M=1k & M=5k & M=10k \\ 
\midrule
\multirow{2}{*}{N/A} & \texttimes & $4.5 \pm 0.6$ & $5.0 \pm 0.4$ & $4.9 \pm 0.2$ & $\bm{10.9 \pm 0.9}$ & $17.5 \pm 1.3$ & $17.1 \pm 1.8$ & $\bm{10.6 \pm 0.5}$ & $\bm{18.4 \pm 1.9}$ & $\bm{17.9 \pm 3.3}$ \\
  & \checkmark & $\bm{5.8 \pm 0.4}$ & $\bm{6.2 \pm 0.2}$ & $\bm{6.2 \pm 0.4}$ & $9.7 \pm 0.6$ & $\bm{18.2 \pm 1.2}$ & $\bm{17.9 \pm 0.9}$ & $8.9 \pm 0.7$ & $16.2 \pm 0.6$ & $16.6 \pm 1.3$ \\
\midrule
\multirow{2}{*}{LB \cite{zeno_task_2019}} & \texttimes & $9.9 \pm 0.9$ & $9.9 \pm 0.6$ & $9.9 \pm 0.6$ & $17.4 \pm 0.8$ & $18.6 \pm 2.2$ & $18.3 \pm 3.1$ & $17.9 \pm 1.3$ & $20.3 \pm 1.1$ & $21.9 \pm 1.2$ \\
  & \checkmark & $\bm{12.2 \pm 0.6}$ & $\bm{12.2 \pm 0.5}$ & $\bm{12.2 \pm 1.0}$ & $\bm{20.5 \pm 0.8}$ & $\bm{24.8 \pm 1.7}$ & $\bm{27.1 \pm 0.9}$ & $\bm{19.5 \pm 1.0}$ & $\bm{25.4 \pm 0.6}$ & $\bm{26.4 \pm 1.2}$ \\
\midrule
\multirow{2}{*}{SS \cite{ahn_ss-il_2022}} & \texttimes & $11.2 \pm 0.5$ & $10.6 \pm 0.9$ & $10.2 \pm 0.9$ & $16.9 \pm 1.1$ & $20.7 \pm 2.3$ & $20.9 \pm 3.2$ & $18.5 \pm 0.5$ & $22.1 \pm 1.5$ & $21.8 \pm 1.5$ \\
  & \checkmark & $\bm{13.8 \pm 1.1}$ & $\bm{13.8 \pm 0.9}$ & $\bm{13.7 \pm 0.6}$ & $\bm{21.0 \pm 0.9}$ & $\bm{28.0 \pm 0.8}$ & $\bm{30.8 \pm 0.8}$ & $\bm{20.8 \pm 0.9}$ & $\bm{26.9 \pm 1.2}$ & $\bm{29.1 \pm 0.5}$ \\
\midrule
\multirow{2}{*}{RV \cite{ferrari_end--end_2018}} & \texttimes & $\bm{8.5 \pm 0.7}$ & $\bm{18.6 \pm 1.3}$ & $\bm{23.7 \pm 1.0}$ & $\bm{14.6 \pm 0.8}$ & $29.3 \pm 1.3$ & $31.5 \pm 1.0$ & $\bm{12.5 \pm 0.6}$ & $\bm{28.8 \pm 1.5}$ & $32.7 \pm 1.4$ \\
  & \checkmark & $6.6 \pm 0.6$ & $13.6 \pm 1.0$ & $23.3 \pm 0.9$ & $11.5 \pm 0.6$ & $\bm{30.1 \pm 0.4}$ & $\bm{33.9 \pm 0.6}$ & $10.2 \pm 0.5$ & $27.4 \pm 0.9$ & $\bm{34.4 \pm 0.8}$ \\
\midrule
\multirow{2}{*}{NCM \cite{rebuffi_icarl_2017}} & \texttimes & $9.9 \pm 0.4$ & $13.0 \pm 0.7$ & $13.4 \pm 0.5$ & $17.6 \pm 0.8$ & $22.7 \pm 1.3$ & $22.6 \pm 1.7$ & $17.6 \pm 0.4$ & $23.2 \pm 1.9$ & $23.7 \pm 2.2$ \\
  & \checkmark & $\bm{14.3 \pm 0.6}$ & $\bm{18.4 \pm 0.7}$ & $\bm{18.7 \pm 0.7}$ & $\bm{18.4 \pm 0.6}$ & $\bm{28.3 \pm 0.6}$ & $\bm{29.5 \pm 0.7}$ & $\bm{18.0 \pm 0.7}$ & $\bm{26.2 \pm 0.7}$ & $\bm{28.5 \pm 0.5}$ \\

\bottomrule
\end{tabular}%
}
\end{table*}

\begin{table*}[t]
\centering
\caption{End Average Accuracy of methods and tricks with or without NGD-KFAC for the OCI setting on CORe50-NC.}
\label{table:core50}
\resizebox{\textwidth}{!}{%
\begin{tabular}{@{}l|c|ccc|ccc|ccc@{}}
\toprule
&Finetune& & \multicolumn{7}{c}{$7.7 \pm 1.0$} &\\
&Offline&  & \multicolumn{7}{c}{$51.7 \pm 1.8$} & \\
\midrule
\multirow{2.5}{*}{Trick} & \multirow{2.5}{*}{NGD-KFAC} & \multicolumn{3}{c}{A-GEM} & \multicolumn{3}{c}{ER} & \multicolumn{3}{c}{MIR} \\ 
\cmidrule(lr){3-11}
 & & M=1k & M=5k  & M=10k & M=1k & M=5k & M=10k & M=1k & M=5k & M=10k \\ 
\midrule
\multirow{2}{*}{N/A} & \texttimes & $9.2 \pm 0.8$ & $8.9 \pm 0.9$ & $8.8 \pm 1.2$ & $23.4 \pm 1.5$ & $27.2 \pm 1.8$ & $28.9 \pm 1.2$ & $24.2 \pm 2.0$ & $31.3 \pm 1.2$ & $31.9 \pm 1.4$ \\
  & \checkmark & $\bm{9.8 \pm 1.2}$ & $\bm{9.4 \pm 1.0}$ & $\bm{9.4 \pm 0.8}$ & $\bm{24.4 \pm 0.8}$ & $\bm{30.8 \pm 1.0}$ & $\bm{32.2 \pm 1.4}$ & $\bm{27.0 \pm 1.0}$ & $\bm{34.2 \pm 1.0}$ & $\bm{35.5 \pm 1.5}$ \\
\midrule
\multirow{2}{*}{LB \cite{zeno_task_2019}} & \texttimes & $13.4 \pm 1.1$ & $13.4 \pm 1.0$ & $\bm{13.6 \pm 1.5}$ & $22.4 \pm 1.2$ & $25.5 \pm 0.9$ & $25.6 \pm 1.8$ & $22.5 \pm 1.4$ & $25.7 \pm 1.2$ & $25.5 \pm 1.6$ \\
  & \checkmark & $\bm{14.2 \pm 1.2}$ & $\bm{14.8 \pm 0.6}$ & $13.5 \pm 0.7$ & $\bm{25.4 \pm 1.1}$ & $\bm{27.8 \pm 1.1}$ & $\bm{29.1 \pm 1.1}$ & $\bm{26.4 \pm 1.2}$ & $\bm{28.2 \pm 1.1}$ & $\bm{28.8 \pm 1.4}$ \\
\midrule
\multirow{2}{*}{SS \cite{ahn_ss-il_2022}} & \texttimes & $13.9 \pm 1.1$ & $14.9 \pm 1.5$ & $\bm{14.9 \pm 1.6}$ & $22.4 \pm 1.2$ & $25.3 \pm 1.0$ & $25.2 \pm 1.6$ & $23.0 \pm 0.8$ & $26.6 \pm 0.8$ & $27.8 \pm 1.4$ \\
  & \checkmark & $\bm{15.4 \pm 1.1}$ & $\bm{15.0 \pm 0.7}$ & $14.5 \pm 1.2$ & $\bm{26.1 \pm 1.2}$ & $\bm{30.5 \pm 1.5}$ & $\bm{30.7 \pm 1.6}$ & $\bm{26.8 \pm 1.5}$ & $\bm{31.3 \pm 1.6}$ & $\bm{31.9 \pm 1.7}$ \\
\midrule
\multirow{2}{*}{RV \cite{ferrari_end--end_2018}} & \texttimes & $\bm{20.8 \pm 1.5}$ & $24.1 \pm 0.8$ & $26.9 \pm 0.9$ & $24.4 \pm 1.1$ & $30.3 \pm 1.3$ & $32.4 \pm 1.1$ & $25.1 \pm 1.4$ & $32.3 \pm 0.9$ & $34.4 \pm 1.3$ \\
  & \checkmark & $15.1 \pm 1.2$ & $\bm{24.2 \pm 1.1}$ & $\bm{30.4 \pm 0.7}$ & $\bm{25.8 \pm 1.1}$ & $\bm{35.1 \pm 1.7}$ & $\bm{37.7 \pm 1.9}$ & $\bm{27.0 \pm 1.1}$ & $\bm{36.7 \pm 1.1}$ & $\bm{38.2 \pm 1.1}$ \\
\midrule
\multirow{2}{*}{NCM \cite{rebuffi_icarl_2017}} & \texttimes & $17.4 \pm 0.9$ & $18.9 \pm 1.0$ & $19.8 \pm 1.1$ & $21.7 \pm 1.3$ & $25.1 \pm 1.8$ & $26.2 \pm 1.2$ & $22.8 \pm 1.1$ & $28.6 \pm 1.1$ & $29.4 \pm 1.4$ \\
  & \checkmark & $\bm{20.4 \pm 1.5}$ & $\bm{21.9 \pm 1.1}$ & $\bm{21.2 \pm 1.3}$ & $\bm{28.5 \pm 1.3}$ & $\bm{34.8 \pm 2.1}$ & $\bm{35.3 \pm 2.4}$ & $\bm{29.5 \pm 1.6}$ & $\bm{34.9 \pm 2.0}$ & $\bm{35.5 \pm 2.4}$ \\

\bottomrule
\end{tabular}%
}
\end{table*}

\subsection{Class incremental}
\subsubsection{Overall Effect of NGD-KFAC}
\begin{itemize}
    \item \textbf{Without NGD-KFAC:} The models achieved an average end accuracy of $19.58\%$ and an average forgetting of $18.05\%$.
    \item \textbf{With NGD-KFAC:} The models' performance improved, with an average end accuracy of $22.20\%$ and an average forgetting of $24.58\%$.
\end{itemize}

This shows the benefit of using NGD-KFAC optimization in OCL, providing an overall increase in model performance by $2.62\%$. However, while NGD-KFAC enhances performance, it slightly increases the tendency of models to forget previously learned information.  

\subsubsection{Effect of Memory Size}
\begin{itemize}
    \item \textbf{Memory Size 1000:} Models without NGD-KFAC had an average end accuracy of $15.52\%$ and forgetting of $23.72\%$, whereas those with NGD-KFAC optimization reached $17.09\%$ in accuracy and $30.05\%$ in forgetting.
    \item \textbf{Memory Size 5000:} Performance increased significantly with memory size, with models achieving $20.95\%$ in accuracy and $15.74\%$ in forgetting without NGD-KFAC, and $23.71\%$ in accuracy and $22.78\%$ in forgetting with NGD-KFAC.
    \item \textbf{Memory Size 10000:} The largest memory size showed the highest performance, with models achieving $22.22\%$ in accuracy and $14.70\%$ in forgetting without NGD-KFAC, and $25.80\%$ in accuracy and $20.44\%$ in forgetting with NGD-KFAC.
\end{itemize}

The results show the importance of memory size in OCL, with larger buffers allowing for better performance. Additionally, the benefit of NGD-KFAC is consistent across different memory sizes, showing larger gains with increased buffer size.

\subsubsection{Effect of OCL Tricks}
The analysis of different OCL tricks reveals varied improvements in performance when combined with NGD-KFAC:

\begin{itemize}
\item \textbf{Labels Trick:} Improved from $17.77\%$ to $20.92\%$ in accuracy with NGD-KFAC, with forgetting increasing from $14.74\%$ to $18.07\%$.
\item \textbf{Nearest Class Mean (NCM) Trick:} Showed a notable increase from $20.89\%$ to $25.31\%$ in accuracy with NGD-KFAC, with forgetting increasing from $11.49\%$ to $12.54\%$.
\item \textbf{Review Trick:} Had a high baseline performance of $24.52\%$ which remained almost the same at $24.91\%$ with NGD-KFAC, but forgetting significantly increased from $17.24\%$ to $31.78\%$.
\item \textbf{Separated Softmax:} Saw an improvement from $18.81\%$ to $22.95\%$ in accuracy with NGD-KFAC, with a modest increase in forgetting from $13.75\%$ to $16.72\%$.
\item \textbf{No Trick:} Here the accuracy went from $15.92\%$ without NGD-KFAC to $16.94\%$ with NGD-KFAC, but with a substantial increase in forgetting from $33.08\%$ to $43.83\%$.
\end{itemize}

These results suggest that NGD-KFAC significantly enhances the effectiveness of various OCL tricks in improving model accuracy, with varied impacts on forgetting. The Nearest Class Mean (NCM) trick, combined with NGD-KFAC is particularly effective, significantly improving performance while reducing forgetting. It is interesting to notice that although the forgetting seems to increase with the use of NGD-KFAC, the accuracy increases with it. This increase in accuracy can be attributed to the faster convergence facilitated by NGD-KFAC in each learning task. As a result, although the initial decrease in performance when switching to a new task is greater compared to using SGD, the overall performance is still better. This might indicate that, with more robust OCL architectures, NGD-KFAC may be able to achieve better performance. We leave further exploration of this hypothesis to future research. 

\begin{figure}[h]
\begin{center}
   \includegraphics[width=\linewidth]{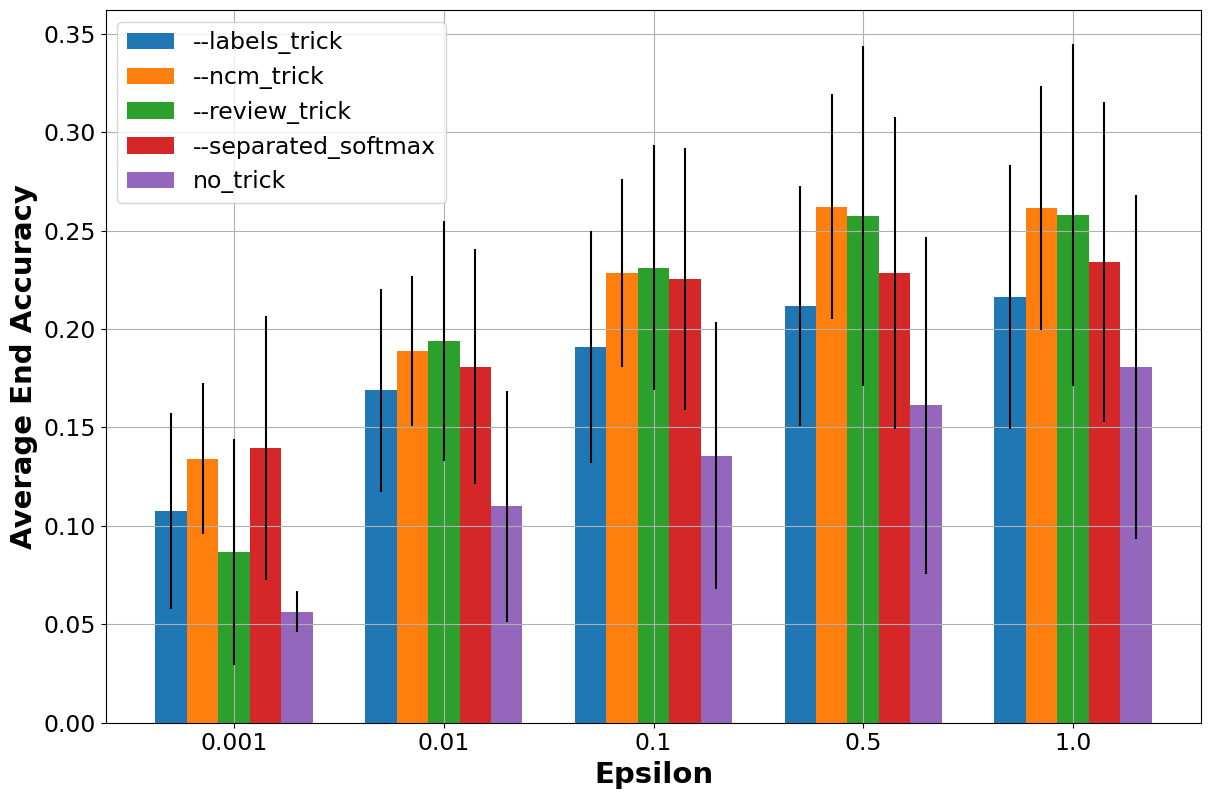}
\end{center}
   \caption{Average End Accuracy by trick, averaged on all datasets in the OCI setting with regards to the damping parameter $\epsilon$.}
\label{fig:reg_grouped_plot_bar}
\end{figure}

\subsection{Domain incremental}

In the domain incremental setting (ODI), the benefit of NGD-KFAC is also clearly visible for most methods and datasets. For example, for experience replay with a memory buffer of 10000, it improves the accuracy from $24.3\%$ to $26.7\%$ in Mini-ImageNet with Noise, from $13.7\%$ to $26.7\%$ in Mini-ImageNet with occlusion, from $9.4\%$ to $10\%$ in Mini-ImageNet with Blur and from $30\%$ to $33.8\%$ in CORe50-NI. 

\subsection{Regularization}

The optimization is quite sensitive to the dampening coefficient $\epsilon$ of the regularisation. We show, in figure \ref{fig:reg_grouped_plot_bar}, the average end accuracy over all datasets in the class incremental setting. Higher dampening coefficients lead to higher accuracy for all the tricks implemented.


\section{Conlusion}

In this paper, we proposed and explored the novel application of Natural Gradient Descent with Kronecker Factored Approximate Curvature (NGD-KFAC) as an optimization method in the domain of Online Continual Learning (OCL) for image classification. Our work demonstrates the advantage of using second-order optimization techniques, specifically NGD-KFAC, to address the challenges presented by OCL, such as catastrophic forgetting and the requirement for rapid convergence despite data non-stationarity and task shifts. Through our experiments across several datasets and OCL scenarios, we established that NGD-KFAC not only improves the performance of OCL models but also enhances the effectiveness of existing continual learning tricks.

Our experiments across various benchmarks, including Split CIFAR-100, CORE50, and Split MiniImageNet, indicate that integrating NGD-KFAC with traditional OCL methods boosts model performance. Specifically, we observed in some cases, improvements in both average accuracy and reduction in average forgetting, showing NGD-KFAC's potential to mitigate the effects of catastrophic forgetting more effectively than traditional first-order optimization approaches. This improvement is due to NGD-KFAC's ability to account for the curvature of the parameter space informed by the data distribution, thereby offering a more principled and efficient update path.


In conclusion, our findings suggest that second-order optimization methods, particularly the Natural Gradient Descent, hold promise for enhancing the performance and robustness of models in dynamic and evolving learning environments. 






\bibliographystyle{plain}
\bibliography{mybibfile}

@article{george_fast_2021,
  title={Fast approximate natural gradient descent in a kronecker factored eigenbasis},
  author={George, Thomas and Laurent, C{\'e}sar and Bouthillier, Xavier and Ballas, Nicolas and Vincent, Pascal},
  journal={Advances in neural information processing systems},
  volume={31},
  year={2018}
}

@inproceedings{he2015deep,
  title={Deep residual learning for image recognition},
  author={He, Kaiming and Zhang, Xiangyu and Ren, Shaoqing and Sun, Jian},
  booktitle={Proceedings of the IEEE conference on computer vision and pattern recognition},
  pages={770--778},
  year={2016}
}

@article{rusu2022progressiveneuralnetworks,
  title={Progressive neural networks},
  author={Rusu, Andrei A and Rabinowitz, Neil C and Desjardins, Guillaume and Soyer, Hubert and Kirkpatrick, James and Kavukcuoglu, Koray and Pascanu, Razvan and Hadsell, Raia},
  journal={arXiv preprint arXiv:1606.04671},
  year={2016}
}

@article{nguyen2018variationalcontinuallearning,
  title={Variational continual learning},
  author={Nguyen, Cuong V and Li, Yingzhen and Bui, Thang D and Turner, Richard E},
  journal={arXiv preprint arXiv:1710.10628},
  year={2017}
}

@inproceedings{aljundi_memory_2018,
  title={Memory aware synapses: Learning what (not) to forget},
  author={Aljundi, Rahaf and Babiloni, Francesca and Elhoseiny, Mohamed and Rohrbach, Marcus and Tuytelaars, Tinne},
  booktitle={Proceedings of the European conference on computer vision (ECCV)},
  pages={139--154},
  year={2018}
}

@article{aljundi2019gradient,
  title={Gradient based sample selection for online continual learning},
  author={Aljundi, Rahaf and Lin, Min and Goujaud, Baptiste and Bengio, Yoshua},
  journal={Advances in neural information processing systems},
  volume={32},
  year={2019}
}

@inproceedings{zenke_continual_2017,
  title={Continual learning through synaptic intelligence},
  author={Zenke, Friedemann and Poole, Ben and Ganguli, Surya},
  booktitle={International conference on machine learning},
  pages={3987--3995},
  year={2017},
  organization={PMLR}
}

@article{soori_tengrad_2022,
  title={TENGraD: Time-efficient natural gradient descent with exact fisher-block inversion},
  author={Soori, Saeed and Can, Bugra and Mu, Baourun and G{\"u}rb{\"u}zbalaban, Mert and Dehnavi, Maryam Mehri},
  journal={arXiv preprint arXiv:2106.03947},
  year={2021}
}

@article{martens_new_2014,
  title={New insights and perspectives on the natural gradient method},
  author={Martens, James},
  journal={Journal of Machine Learning Research},
  volume={21},
  number={146},
  pages={1--76},
  year={2020}
}

@article{shrestha_natural_2023,
  title={Natural gradient methods: Perspectives, efficient-scalable approximations, and analysis},
  author={Shrestha, Rajesh},
  journal={arXiv preprint arXiv:2303.05473},
  year={2023}
}

@article{heskes_natural_2000,
  title={On “natural” learning and pruning in multilayered perceptrons},
  author={Heskes, Tom},
  journal={Neural Computation},
  volume={12},
  number={4},
  pages={881--901},
  year={2000},
  publisher={MIT Press}
}

@article{desjardins_natural_2015,
  title={Natural neural networks},
  author={Desjardins, Guillaume and Simonyan, Karen and Pascanu, Razvan and others},
  journal={Advances in neural information processing systems},
  volume={28},
  year={2015}
}

@article{amari_natural_1998,
  title={Natural gradient works efficiently in learning},
  author={Amari, Shun-Ichi},
  journal={Neural computation},
  volume={10},
  number={2},
  pages={251--276},
  year={1998},
  publisher={MIT Press}
}

@inproceedings{martens_optimizing_2020,
  title={Optimizing neural networks with kronecker-factored approximate curvature},
  author={Martens, James and Grosse, Roger},
  booktitle={International conference on machine learning},
  pages={2408--2417},
  year={2015},
  organization={PMLR}
}

@inproceedings{fujimoto_neural_2017,
  title={A neural network model with bidirectional whitening},
  author={Fujimoto, Yuki and Ohira, Toru},
  booktitle={Artificial Intelligence and Soft Computing: 17th International Conference, ICAISC 2018, Zakopane, Poland, June 3-7, 2018, Proceedings, Part I 17},
  pages={47--57},
  year={2018},
  organization={Springer}
}

@article{van_de_ven_three_2022,
  title={Three types of incremental learning},
  author={Van de Ven, Gido M and Tuytelaars, Tinne and Tolias, Andreas S},
  journal={Nature Machine Intelligence},
  volume={4},
  number={12},
  pages={1185--1197},
  year={2022},
  publisher={Nature Publishing Group UK London}
}

@article{rolnick_experience_2019,
  title={Experience replay for continual learning},
  author={Rolnick, David and Ahuja, Arun and Schwarz, Jonathan and Lillicrap, Timothy and Wayne, Gregory},
  journal={Advances in neural information processing systems},
  volume={32},
  year={2019}
}

@article{hu2020gradient,
  title={Gradient episodic memory with a soft constraint for continual learning},
  author={Hu, Guannan and Zhang, Wu and Ding, Hu and Zhu, Wenhao},
  journal={arXiv preprint arXiv:2011.07801},
  year={2020}
}

@article{lopez-paz_gradient_2022,
  title={Gradient episodic memory for continual learning},
  author={Lopez-Paz, David and Ranzato, Marc'Aurelio},
  journal={Advances in neural information processing systems},
  volume={30},
  year={2017}
}

@article{french_catastrophic_1999,
  title={Catastrophic forgetting in connectionist networks},
  author={French, Robert M},
  journal={Trends in cognitive sciences},
  volume={3},
  number={4},
  pages={128--135},
  year={1999},
  publisher={Elsevier}
}

@incollection{mccloskey_catastrophic_1989,
  title={Catastrophic interference in connectionist networks: The sequential learning problem},
  author={McCloskey, Michael and Cohen, Neal J},
  booktitle={Psychology of learning and motivation},
  volume={24},
  pages={109--165},
  year={1989},
  publisher={Elsevier}
}

@article{kudithipudi_biological_2022,
  title={Biological underpinnings for lifelong learning machines},
  author={Kudithipudi, Dhireesha and Aguilar-Simon, Mario and Babb, Jonathan and Bazhenov, Maxim and Blackiston, Douglas and Bongard, Josh and Brna, Andrew P and Chakravarthi Raja, Suraj and Cheney, Nick and Clune, Jeff and others},
  journal={Nature Machine Intelligence},
  volume={4},
  number={3},
  pages={196--210},
  year={2022},
  publisher={Nature Publishing Group UK London}
}

@article{li_learning_2017,
  title={Learning without forgetting},
  author={Li, Zhizhong and Hoiem, Derek},
  journal={IEEE transactions on pattern analysis and machine intelligence},
  volume={40},
  number={12},
  pages={2935--2947},
  year={2017},
  publisher={IEEE}
}

@article{kirkpatrick_overcoming_2017,
  title={Overcoming catastrophic forgetting in neural networks},
  author={Kirkpatrick, James and Pascanu, Razvan and Rabinowitz, Neil and Veness, Joel and Desjardins, Guillaume and Rusu, Andrei A and Milan, Kieran and Quan, John and Ramalho, Tiago and Grabska-Barwinska, Agnieszka and others},
  journal={Proceedings of the national academy of sciences},
  volume={114},
  number={13},
  pages={3521--3526},
  year={2017},
  publisher={National Academy of Sciences}
}

@inproceedings{ferrari_end--end_2018,
  title={End-to-end incremental learning},
  author={Castro, Francisco M and Mar{\'\i}n-Jim{\'e}nez, Manuel J and Guil, Nicol{\'a}s and Schmid, Cordelia and Alahari, Karteek},
  booktitle={Proceedings of the European conference on computer vision (ECCV)},
  pages={233--248},
  year={2018}
}

@inproceedings{rebuffi_icarl_2017,
  title={icarl: Incremental classifier and representation learning},
  author={Rebuffi, Sylvestre-Alvise and Kolesnikov, Alexander and Sperl, Georg and Lampert, Christoph H},
  booktitle={Proceedings of the IEEE conference on Computer Vision and Pattern Recognition},
  pages={2001--2010},
  year={2017}
}

@article{mai_online_2022,
  title={Online continual learning in image classification: An empirical survey},
  author={Mai, Zheda and Li, Ruiwen and Jeong, Jihwan and Quispe, David and Kim, Hyunwoo and Sanner, Scott},
  journal={Neurocomputing},
  volume={469},
  pages={28--51},
  year={2022},
  publisher={Elsevier}
}

@article{zeno_task_2019,
  title={Task-agnostic continual learning using online variational bayes with fixed-point updates},
  author={Zeno, Chen and Golan, Itay and Hoffer, Elad and Soudry, Daniel},
  journal={Neural Computation},
  volume={33},
  number={11},
  pages={3139--3177},
  year={2021},
  publisher={MIT Press One Rogers Street, Cambridge, MA 02142-1209, USA journals-info~…}
}

@inproceedings{tseran2018natural,
  title={Natural variational continual learning},
  author={Tseran, Hanna and Khan, Mohammad Emtiyaz and Harada, Tatsuya and Bui, Thang D},
  booktitle={Continual Learning Workshop@ NeurIPS},
  volume={2},
  year={2018}
}

@article{aljundi_online_2019,
  title={Online continual learning with maximal interfered retrieval},
  author={Aljundi, Rahaf and Belilovsky, Eugene and Tuytelaars, Tinne and Charlin, Laurent and Caccia, Massimo and Lin, Min and Page-Caccia, Lucas},
  journal={Advances in neural information processing systems},
  volume={32},
  year={2019}
}

@inproceedings{ahn_ss-il_2022,
  title={Ss-il: Separated softmax for incremental learning},
  author={Ahn, Hongjoon and Kwak, Jihwan and Lim, Subin and Bang, Hyeonsu and Kim, Hyojun and Moon, Taesup},
  booktitle={Proceedings of the IEEE/CVF International conference on computer vision},
  pages={844--853},
  year={2021}
}

@article{masana_class-incremental_2022,
  title={Class-incremental learning: survey and performance evaluation on image classification},
  author={Masana, Marc and Liu, Xialei and Twardowski, Bart{\l}omiej and Menta, Mikel and Bagdanov, Andrew D and Van De Weijer, Joost},
  journal={IEEE Transactions on Pattern Analysis and Machine Intelligence},
  volume={45},
  number={5},
  pages={5513--5533},
  year={2022},
  publisher={IEEE}
}

@inproceedings{hou_learning_2019,
  title={Learning a unified classifier incrementally via rebalancing},
  author={Hou, Saihui and Pan, Xinyu and Loy, Chen Change and Wang, Zilei and Lin, Dahua},
  booktitle={Proceedings of the IEEE/CVF conference on computer vision and pattern recognition},
  pages={831--839},
  year={2019}
}

@inproceedings{wu_large_2019,
  title={Large scale incremental learning},
  author={Wu, Yue and Chen, Yinpeng and Wang, Lijuan and Ye, Yuancheng and Liu, Zicheng and Guo, Yandong and Fu, Yun},
  booktitle={Proceedings of the IEEE/CVF conference on computer vision and pattern recognition},
  pages={374--382},
  year={2019}
}

@inproceedings{ba_distributed_2016,
  title={Distributed second-order optimization using kronecker-factored approximations},
  author={Ba, Jimmy and Grosse, Roger and Martens, James},
  booktitle={International conference on learning representations},
  year={2017}
}

@article{vinyals2017matching,
  title={Matching networks for one shot learning},
  author={Vinyals, Oriol and Blundell, Charles and Lillicrap, Timothy and Wierstra, Daan and others},
  journal={Advances in neural information processing systems},
  volume={29},
  year={2016}
}

@inproceedings{More1977TheLA,
  title={The Levenberg-Marquardt algorithm: implementation and theory},
  author={Mor{\'e}, Jorge J},
  booktitle={Numerical analysis: proceedings of the biennial Conference held at Dundee, June 28--July 1, 1977},
  pages={105--116},
  year={2006},
  organization={Springer}
}

@article{Darken1990NoteOL,
  title={Note on learning rate schedules for stochastic optimization},
  author={Darken, Christian and Moody, John},
  journal={Advances in neural information processing systems},
  volume={3},
  year={1990}
}

@article{goldfarb2021practical,
  title={Practical quasi-newton methods for training deep neural networks},
  author={Goldfarb, Donald and Ren, Yi and Bahamou, Achraf},
  journal={Advances in Neural Information Processing Systems},
  volume={33},
  pages={2386--2396},
  year={2020}
}

@inproceedings{lomonaco_core50_2017,
  title={Core50: a new dataset and benchmark for continuous object recognition},
  author={Lomonaco, Vincenzo and Maltoni, Davide},
  booktitle={Conference on robot learning},
  pages={17--26},
  year={2017},
  organization={PMLR}
}

@article{krizhevsky_learning_2009,
  title={Learning multiple layers of features from tiny images},
  author={Krizhevsky, Alex and Hinton, Geoffrey and others},
  year={2009},
  publisher={Toronto, ON, Canada}
}

@article{chaudhry2019tiny,
  title={On tiny episodic memories in continual learning},
  author={Chaudhry, Arslan and Rohrbach, Marcus and Elhoseiny, Mohamed and Ajanthan, Thalaiyasingam and Dokania, Puneet K and Torr, Philip HS and Ranzato, Marc'Aurelio},
  journal={arXiv preprint arXiv:1902.10486},
  year={2019}
}

@inproceedings{soutif--cormerais_comprehensive_2023,
  title={A comprehensive empirical evaluation on online continual learning},
  author={Soutif-Cormerais, Albin and Carta, Antonio and Cossu, Andrea and Hurtado, Julio and Lomonaco, Vincenzo and Van de Weijer, Joost and Hemati, Hamed},
  booktitle={Proceedings of the IEEE/CVF International Conference on Computer Vision},
  pages={3518--3528},
  year={2023}
}

\end{document}